\documentclass{article}
\usepackage{amssymb}
\usepackage{booktabs}
\usepackage{amssymb}
\usepackage{graphicx}
\usepackage{amsmath}
\usepackage{wrapfig}
\usepackage{caption}
\usepackage{color}
\usepackage{makecell}
\usepackage{pifont} 
\usepackage{tikz}   
\usepackage{bbm}

\usepackage[preprint]{corl_2025} 
\usepackage{amssymb}
\usepackage{enumitem}

\title{
Learning to Plan \& Schedule with 
Reinforcement-Learned Bimanual Robot Skills
}

\author{
  Weikang Wan\textsuperscript{1,2}\thanks{Work done while interning at NVIDIA Research.}\quad
  Fabio Ramos\textsuperscript{1}\quad
  Xuning Yang\textsuperscript{1}\quad
  Caelan Garrett\textsuperscript{1}\\[0.5ex]
  \normalfont
  \textsuperscript{1}NVIDIA \quad
  \textsuperscript{2}University of California San Diego \quad 
  \\
  \texttt{{\small \{weikangw,ftozetoramos,xuningy,cgarrett\}@nvidia.com}} \\
}

\usepackage{algorithm}
\usepackage[noend]{algpseudocode}
\usepackage{amssymb}
\usepackage{amsmath}
\usepackage[normalem]{ulem}
\usepackage{wrapfig}
\usepackage{bm}

\usepackage{amsthm}
\theoremstyle{plain}
\theoremstyle{definition}
\theoremstyle{plain}
\theoremstyle{plain}

\usepackage{listings}
\begin{document}
\maketitle


\begin{abstract}
    Long-horizon contact-rich bimanual manipulation presents a significant challenge, requiring complex coordination involving a mixture of parallel execution and sequential collaboration between arms. 
    In this paper, we introduce a hierarchical framework that frames this challenge as an integrated skill planning \& {\em scheduling} problem, going beyond purely sequential decision-making to support simultaneous skill invocation.
    Our approach is built upon a library of single-arm and bimanual primitive skills, each trained using Reinforcement Learning (RL) in GPU-accelerated simulation.
    We then train a Transformer on a dataset of skill compositions to act as a high-level planner \& scheduler, simultaneously predicting the discrete schedule of skills as well as their continuous parameters. 
    We demonstrate that our method achieves higher success rates on complex, contact-rich tasks than end-to-end RL approaches and produces more efficient, coordinated behaviors than traditional sequential-only planners.
\end{abstract}




\section{Introduction}
Enabling robots to perform long-horizon, contact-rich manipulation is a longstanding goal in robotics, with bimanual systems offering the potential for human-like dexterity~\citep{smith2012dual,shome2021fast}. 
A primary challenge in this domain lies in programming two arms to act in harmony to accomplish a complex goal. 
Such coordination demands a flexible control strategy that combines parallel, serial, and collaborative execution of skills. 

Many previous approaches model manipulation as a sequential decision-making process, where a policy selects a single action or skill at each step~\citep{zhu2022bottom,nasiriany2022augmenting,jiang2024hacman++}. 
This formulation, however, creates an inherent bottleneck for bimanual tasks, as it fails to capture opportunities for simultaneous execution and can lead to inefficient, underutilized behaviors. 
In this paper, we argue that the high-level decision problem is better framed as an integrated planning and scheduling problem, where the goal is to assign tasks to both arms, where at some points in time the arms act independently and at other points they act collaboratively.

To this end, we propose a hierarchical framework built upon a library of reinforcement-learned primitive skills. 
Our Transformer-based high-level policy functions as a skill scheduler, generating plans that specify the discrete skills and their continuous parameters for both arms.
We highlight three contributions of this work:
\begin{itemize}
    \item We propose a novel approach for bimanual manipulation that learns a library of arm-specific primitive skills using Reinforcement Learning (RL) and then through integrated skill planning \& {\em scheduling} combines these in serial and parallel over time.
    \item We show how a Transformer-based scheduling policy can be trained to generate bimanual schedules that specify both discrete skills and their continuous parameters.
    \item We demonstrate through experiments that our approach achieves significantly higher success rates and efficiency compared to end-to-end and sequential planning baselines.
\end{itemize}
\section{Related Work}
\noindent \textbf{Task and Motion Planning.} 
Task and Motion Planning (TAMP)~\citep{garrett2021integrated,kaelbling2011hierarchical,dantam2016incremental,toussaint2018differentiable,garrett2020pddlstream,shen2024cutamp} algorithms plan sequences of high-level actions and corresponding low-level motion plans, typically for single-robot scenarios.
Multi-robot manipulation and TAMP~\citep{koga1994multi,harada2014manipulation,huang2025apex} extend these principles to plan for multiple robots or arms and introduce opportunities for parallelism, 
but the jump to multiple robots and introduces significant complexity in coordination, collision avoidance, and task scheduling.
Classical TAMP approaches rely on predefined models and constraints, which can be difficult to engineer for tasks requiring non-prehensile manipulation.
Several recent learning for TAMP approaches relax this assumption by replacing handcrafted model components with learned ones~\citep{simeonov2021long,liang2022learning,zhu2021hierarchical,jiang2024hacman++}. 
Others use imitation learning to plan~\cite{yang2023piginet,dalal2023imitating} and implicitly learn a planning model from demonstrations.
Our approach also uses imitation learning for high-level reasoning but predicts bimanual behaviors, scheduling which skills should be used and when, both in serial and in parallel.

\noindent \textbf{Hierarchical Modeling in Robotic Manipulation. }
Hierarchical manipulation methods~\citep{wang2021learning,hedegaard2025beyond} build complex manipulation behaviors from a set of learned primitive control policies.
Recent works use imitation~\citep{pastor2009learning,xu2018neural,mandlekar2023humanintheloop} or reinforcement learning~\citep{sutton1999between,kulkarni2016hierarchical,zhou2024spire} to acquire these 
primitive policies, adding flexibility while benefiting from the temporal abstraction of primitives.
However, these methods are often limited to single-arm scenarios and confined to prehensile manipulation primitives.
In contrast, our method employs a set of versatile parameterized contact-rich single-arm and bimanual primitives for bimanual manipulation tasks.

\begin{figure}[t]
    \centering
    \includegraphics[width=\textwidth]{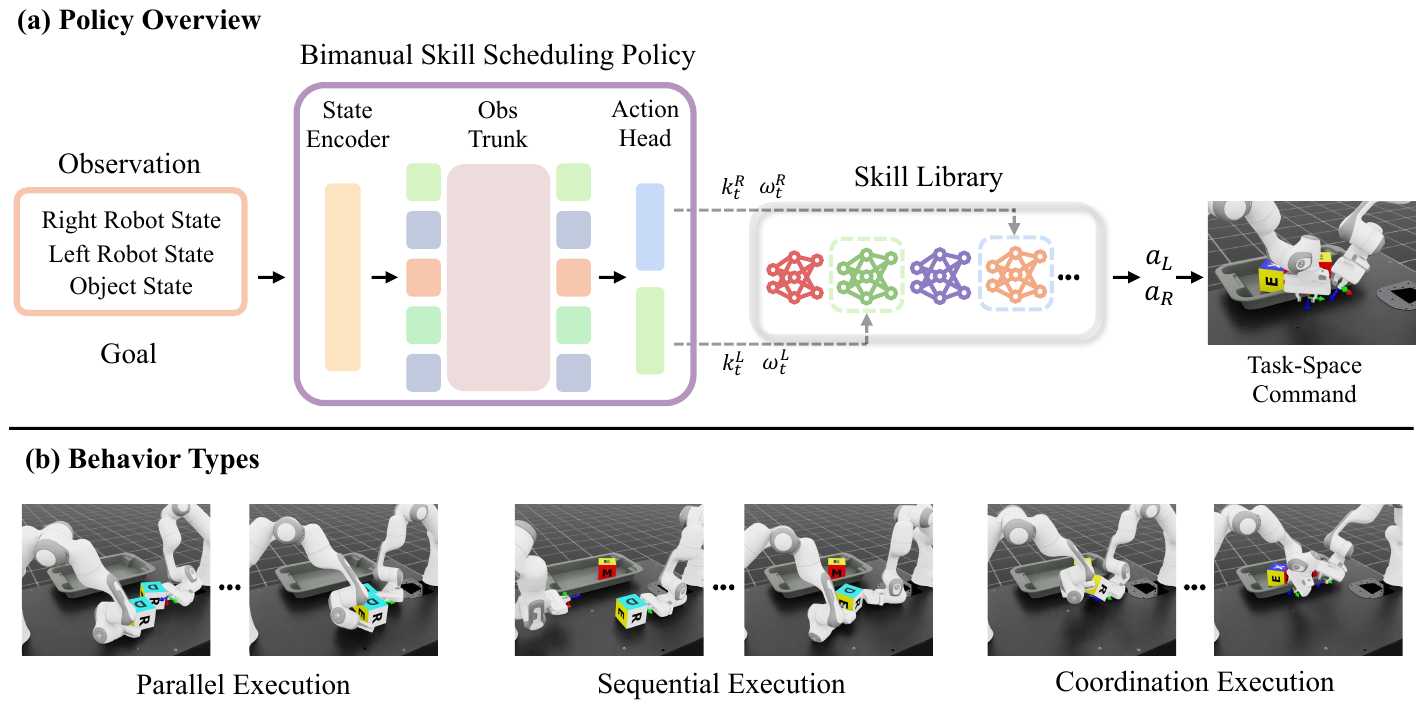}
    \caption{\textbf{(a) Policy Overview.} In our hierarchical framework, the high-level policy selects skills by predicting the skill index $k^R, k^L$ and specifying the skill parameters $\omega^R, \omega^L$ for both arms. \textbf{(b) Behavior Types.} Our method exhibits diverse bimanual behaviors in long-horizon tasks.}
    \label{fig:pipeline}
    \vspace{-4mm}
\end{figure}

\section{Method}


An overview of our pipeline is shown in Fig.~\ref{fig:pipeline}. 
We first define the problem formulation in Sec.~\ref{sec:problem}. 
We then introduce the details of the individual low-level skill training and the high-level bimanual skill scheduling policy in Sec.~\ref{sec:low} and Sec.~\ref{sec:high}.

\subsection{Problem Formulation}
\label{sec:problem}
We formulate the long-horizon bimanual manipulation task as a two-level hierarchical decision problem. We first define the low-level MDP for learning individual skills and then describe the high-level MDP for scheduling these skills to achieve a final task goal $g \in \mathcal{G}$.
Each primitive skill $k$ from a finite library $\mathcal{K}$ is learned as a low-level policy $\pi^{\mathrm{L}}_{k}(a_t \mid s_t,\omega_t)$. This policy solves a short-horizon, goal-conditioned Markov Decision Process (MDP) defined by the tuple $\langle \mathcal{S},\mathcal{A},\mathcal{T},\mathcal{R}_k,\gamma,\Omega\rangle$.
Here, $s_t \in \mathcal{S}$ represents the state at time $t$, $a_t \in \mathcal{A}$ is the low-level action, $\mathcal{T}$ is the state transition function, and $\gamma \in [0, 1)$ is the discount factor.
The goal for this MDP is a skill parameter $\omega_t \in \Omega$ (e.g., a target object pose), and the policy is rewarded by a dense reward function $\mathcal{R}_k(s_t, a_t, \omega_t)$ for making progress towards this subgoal $\omega_t$.
Building on the skill library, the high-level problem is to select the skill $k_t$ and its parameters $\omega_t$ to achieve the overall task goal $g$. This is governed by a high-level scheduling policy $\pi^{\mathrm{H}}(k_t, \omega_t \mid s_t, g)$. 
This high-level MDP is related to a semi-MDP~\cite{sutton1999between}, where each low-level skill acts as a temporally-extended action, similar to that of an option~\cite{stolle2002learning}. 
Each high-level ``action" $(k_t, \omega_t)$ is executed by the corresponding low-level policy $\pi^{\mathrm{L}}_{k_t}$ for multiple timesteps until termination.
The final, factorized policy for the complete task is thus: $ \pi(a_t \mid s_t,g) = \pi^{\mathrm{H}}(k_t, \omega_t \mid s_t,g)\;\pi^{\mathrm{L}}_{k_t}(a_t \mid s_t,\omega_t) $.




\subsection{Individual Low-Level Skills}
\label{sec:low}


We train a set of primitive single-arm and bimanual skills $\pi^{\mathrm{L}}_{k}(a_t \mid s_t,\omega_t)$ using goal-conditioned reinforcement learning.
Each low-level skill policy takes as input the current state $s_t$ and its parameters $\omega_t$. For a single arm, $s_t$ is comprised of the arm’s proprioception (including joint positions, velocities, and end-effector pose), other arm’s end-effector pose, and the object state. The parameters $\omega_t$ encode task-specific information such as the object goal pose. The policy outputs a target end-effector pose action $a_t$, which is then converted to joint torques via an Operational Space Controller (OSC)~\cite{khatib1987unified}.

The low-level primitive skills are trained with dense rewards that combine a contact reward, task-relevant goal reaching reward, a binary success bonus, and an energy penalty. We use a unified template \(r_t = r_{\text{contact}} + r_{\text{goal}} + r_{\text{suc}} - c_{\text{energy}}\), where the task success reward \(r_{\text{suc}}=\mathbf{1}_{\text{suc}}\) is given when the pose of the object is within \(0.05\,\mathrm{m}\) and \(0.1\) radians of the target pose. We add dense shaping \(r_{\text{contact}}=\alpha_c\!\left(1-\tanh(d_c/\sigma_c)\right)\) and \(r_{\text{goal}}=r_{\text{pos}}+r_{\text{rot}}=\alpha_p\!\left(1-\tanh(d_p/\sigma_p)\right)+\alpha_r\!\left(1-\tanh(d_r/\sigma_r)\right)\), where \(d_c\) is the right end-effector–to–object distance, \(d_p\) is the object–to–goal position error, and \(d_r\) is the object–to–goal rotation error; \(\alpha_c,\alpha_p,\alpha_r>0\) are scaling coefficients and \(\sigma_c{=}0.2,\ \sigma_p{=}0.05,\ \sigma_r{=}10\)
are shaping scales; \(c_{\text{energy}}\) denotes an energy penalty: $c_{\text{energy}} = c_e \sum_{i=1}^{J} \tau_i \, \dot{q}_i $, where $c_e \in \mathbb{R}^+$ is a scaling coefficient, and $\tau_i$ and $\dot{q}_i$ are the joint torque and velocity of the $i^\text{th}$ joint.

In this work, we focus on five primitive skills: single-arm pushing, single-arm rotating, bimanual rotating, bimanual pushing, and bimanual pick-and-place. 
These skills were selected by human modelers to cover the fundamental capabilities required by our tasks.
For all skills, the goal is specified by an object target position, an orientation, or full pose. 
Our approach is not limited to these skills and can in principle be applied to more than two robots.

\subsection{Bimanual Skill Scheduling Policy}
\label{sec:high}
After the primitive skills are trained, they can be composed to complete difficult long-horizon tasks. 
We learn the high-level bimanual skill scheduling policy $\pi^{\mathrm{H}}$ through behavior cloning~\citep{pomerleau1988alvinn}. 
The goal is to train a policy that mimics the output of an expert planner that has access to privileged information used to generate the problem.
The learned policy $\pi^{\mathrm{H}}(\omega_t,k_t \mid s_t,g)$ takes in the current observation and produces two types of outputs: a discrete skill $k_t$ for each arm and the continuous skill parameters $\omega_t$ for the selected skill, which in this work, is a goal object pose.

Given the task goal $g$ and current state information $s_t$, the high-level skill scheduling policy decides which skill and which $\omega_t$ to execute for each arm. To understand the task progress, $\pi^{\mathrm{H}}$ perceives the current object state, the task goal, and the proprioception states of both arms. 
The skill scheduling policy utilizes a transformer-based architecture that takes a sequence of recent observations as input, which aims to improve the temporal consistency of the policy's predictions. 
As an example, for the task of placing two tabletop objects into a grey bin in Fig.~\ref{fig:task}, the skill scheduling policy first calls pushing skills for both arms at the same time to move each object toward a suitable central position, and then sequentially invokes bimanual rotating and bimanual pick-and-place skills on the two objects with appropriate skill parameters to complete the task.

\subsection{Data Generation. }
To train the skill scheduling policy, we first generate a dataset of expert demonstrations. Specifically, we implement a custom demonstration generator that has access to privileged problem information. This generator programmatically constructs multiple successful skill sequences, each being an ordered composition of low-level skills. For each sequence, it samples skill parameters $\omega$ and executes the full plan, retaining only successful rollouts. 
During data generation, we additionally define a single-arm \emph{waiting} skill that keeps one arm stationary. For example, for the bimanual rotating skill we sample the object goal rotation within a specified range. This data generation method yields diverse successful sequence data, covering different ways of bimanual division of labor, asynchronous execution, and collaboration.

\subsection{Policy Learning. }
Once we have collected the required skill-sequence data, we train a Transformer-based bimanual skill scheduling policy $\pi^{\mathrm{H}}(\omega_t,k_t \mid s_t,g)$. The training supervision for $\pi^{\mathrm{H}}$ comes from the skill-index labels and skill parameters for both arms produced by the data-generation pipeline. We use a cross-entropy loss for skill-index prediction and an Mean Squared Error (MSE) loss for skill-parameter regression; the objective is \( \mathcal{L}=\mathbb{E}_t\!\big[\sum_{u\in\{L,R\}} \mathrm{CE}(k_{t,u}^{\ast}, \hat{k}_{t,u}) + \lambda_{\omega}\,\|\omega_{t,u}^{\ast}-\hat{\omega}_{t,u}\|_2^2 \big] \), where $\lambda_{\omega}$ balances the two terms.


\begin{figure}[t]
    \centering
    \includegraphics[width=\textwidth]{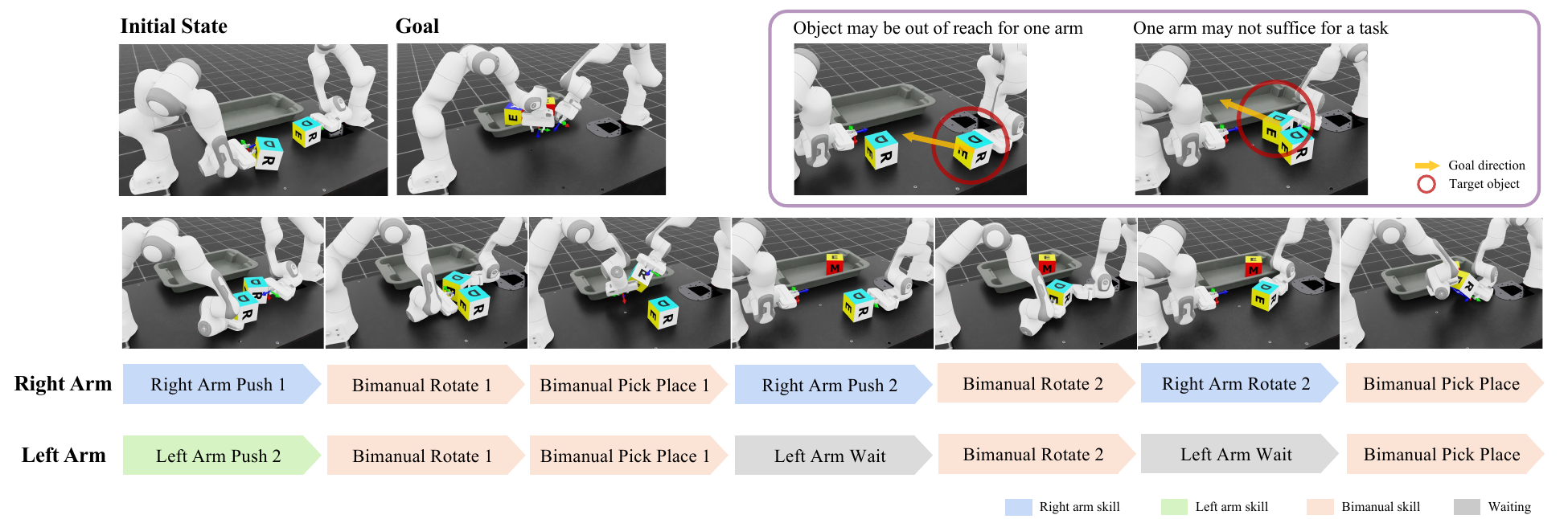}
    \caption{\textbf{ Evaluation Visualization. } The long-horizon rearranging task presents diverse scenarios, such as objects being out of reach or requiring bimanual effort. Our policy handles these by planning the division of labor and collaboration between arms and correctly sequencing the necessary skills.
    }
    \label{fig:task}
    \vspace{-4mm}
\end{figure}

\section{Experiments}
We evaluate our approach with simulated experiments to answer the following questions:
\textbf{Q1.} Does our hierarchical framework achieve higher success rates on long-horizon bimanual manipulation tasks compared to flat end-to-end approaches?
\textbf{Q2.} Can our bimanual skill scheduling policy effectively generate valid schedules that sequence both discrete skills and their continuous parameters for two arms simultaneously?  
\textbf{Q3.} Does our planing \& scheduling formulation lead to more efficient and coordinated bimanual behaviors?

\subsection{Experiments Setup}

We evaluate our framework on a long-horizon, contact-rich manipulation task with goal that: ``\textit{the objects are in the bin}". This task involves placing one or two bulky objects into a bin, where the objects are too large for a single arm to grasp, lift, or reorient alone, which necessitates non-prehensile manipulation.
Furthermore, an object's initial position may be out of reach for one arm. Consequently, this task demands a dynamic combination of strategies: from synchronized coordination for collaborative lifting to independent, asynchronous execution for repositioning. The variability in initial object layouts requires the policy to plan and schedule different skill sequences for each scenario. We build this environment and conduct all simulation experiments in IsaacLab~\citep{mittal2023orbit,NVIDIA_Isaac_Sim}. To compare different approaches, we use three metrics: task success rate (SR) and task completion progress (CP), which is defined as the percentage of optimally-ordered task stages completed, and episode duration (ED) to measure completion efficiency.

\subsection{Quantitative Results}
\noindent \textbf{Baselines.}
We compare our proposed framework against several representative baselines and ablations. We consider the following baselines:
1) RL-scratch is vanilla PPO~\citep{schulman2017proximal} algorithm which learns the entire task from scratch. 
2) Hierarchical RL (HRL)~\citep{kulkarni2016hierarchical} is a hierarchical approach where a high-level policy selects the pre-trained low-level skills using RL. 
3) Sequential-only planning~\citep{simeonov2021long} is a baseline that can only select one skill at each decision step, thus enforcing sequential, non-parallel execution.
4) Ours (Single Arm Ablation) is an ablation of our method constrained to use only a single arm.
For all hierarchical methods, we use the same library of pre-trained low-level skills.

\begin{table}[h!]
\centering
\small 
\begin{tabular}{@{}lrrrrrr@{}}
\toprule
\multicolumn{1}{c}{\textbf{Method}} & \multicolumn{3}{c}{\textbf{Rearrange One Object}} & \multicolumn{3}{c}{\textbf{Rearrange Two Objects}} \\ 
\cmidrule(lr){2-4} \cmidrule(lr){5-7}
 & SR (\%) $\uparrow$ & CP (\%) $\uparrow$ & ED (s) $\downarrow$ & SR (\%) $\uparrow$ & CP (\%) $\uparrow$ & ED (s) $\downarrow$ \\ \midrule
RL-Scratch & $0.0 \pm 0.0$ & $26.4 \pm 2.2$ & $10.0 \pm 0.0$ & $0.0 \pm 0.0$ & $12.3 \pm 1.7$ & $20.0 \pm 0.0$ \\
Hierarchical RL & $20.0 \pm 5.5$ & $42.2 \pm 6.1$ & $9.1 \pm 0.3$ & $7.4 \pm 3.2$ & $19.8 \pm 2.0$ & $19.3 \pm 0.3$ \\ 
Sequential Planning & $48.2 \pm 3.0$ & $\mathbf{66.5 \pm 3.7}$ & $7.6 \pm 0.2$ & $32.4 \pm 3.2$ & $47.9 \pm 3.2$ & $17.1 \pm 0.2$ \\
\midrule
Ours (Single Arm) & $0.0 \pm 0.0$ & $32.1 \pm 2.5$ & $10.0 \pm 0.0$ & $0.0 \pm 0.0$ & $14.0 \pm 1.6$ & $20.0 \pm 0.0$ \\
\textbf{Ours} & $\mathbf{51.3 \pm 2.0}$ & $65.2 \pm 3.8$ & $\mathbf{6.4 \pm 0.2}$ & $\mathbf{38.7 \pm 2.4}$ & $\mathbf{56.1 \pm 4.8}$ & $\mathbf{14.3 \pm 0.2}$ \\ \bottomrule
\end{tabular}
\vspace{2mm}
\caption{Comparison of our method against baselines. We report the mean and standard deviation of 
 the Success Rate (SR), Completion Progress (CP), and Episode Duration (ED) over 100 rollouts.}
\label{tab:main_results}
\end{table}

\noindent \textbf{Results and analysis.}
Table~\ref{tab:main_results} provides comprehensive results for all methods on two task variations: 1) rearranging one object and 2) rearranging two objects into a bin. It answers \textbf{Q1} and \textbf{Q2} by showing that our method outperforms the RL-scratch baseline, achieving a 45\% higher Success Rate (SR) and 41\% higher Task Completion Progress (CP).
Additionally, we visualize the bimanual skill schedule during evaluation in Fig.~\ref{fig:task}, showing that our bimanual skill scheduling policy not only selects the correct single-arm or bimanual skills in different situations, but also efficiently sequences skills for both arms to complete the task. We answer \textbf{Q3} by comparing our method with Sequential Planning, which results in a 16\% reduction in Episode Duration (ED) and highlights our method's ability to more efficiently plan and schedule for both arms simultaneously.
Policy rollout videos are available
at 
\url{https://www.youtube.com/watch?v=zBxGXFzRgGA}.
\section{Conclusion}
In this work, we introduced a hierarchical framework that addresses long-horizon bimanual manipulation by formulating it as an integrated skill planning \& scheduling problem. Our method utilizes a Transformer-based policy to generate coordinated plans, simultaneously selecting discrete skills and regressing their continuous parameters. Experimental results validate that our approach achieves much higher success rates than end-to-end RL and produces more efficient, parallelized behaviors than planners restricted to sequential actions.

\clearpage

\bibliography{ref}  



\clearpage


\end{document}